\documentclass[compsoc,conference,a4paper,10pt,times]{IEEEtran}
\IEEEoverridecommandlockouts
\usepackage{cite}
\usepackage{amsmath,amssymb,amsfonts}
\usepackage{graphicx}
\usepackage{textcomp}
\usepackage{bmpsize}
\usepackage{xcolor}
\usepackage{lipsum}
\usepackage{bbm}
\usepackage{algpseudocode}
\usepackage{algorithm}
\usepackage{float}

\usepackage[colorlinks=true,urlcolor=black]{hyperref}
\def\BibTeX{{\rm B\kern-.05em{\sc i\kern-.025em b}\kern-.08em
	T\kern-.1667em\lower.7ex\hbox{E}\kern-.125emX}}
\begin{document}
 
\title{Learning to Defend by Attacking (and Vice-Versa): \\ Transfer of Learning in Cybersecurity Games}

\author{\IEEEauthorblockN{Tyler Malloy}
\IEEEauthorblockA{Social and Decision Sciences Department\\
Carnegie Mellon University\\
tylerjmalloy@cmu.edu}
\and
\IEEEauthorblockN{Cleotilde Gonzalez}
\IEEEauthorblockA{Social and Decision Sciences Department\\
Carnegie Mellon University\\
coty@cmu.edu}}
\maketitle

\begin{abstract}
Designing cyber defense systems to account for cognitive biases in human decision making has demonstrated significant success in improving performance against human attackers. However, much of the attention in this area has focused on relatively simple accounts of biases in human attackers, and little is known about adversarial behavior or how defenses could be improved by disrupting attacker's behavior. In this work, we present a novel model of human decision-making inspired by the cognitive faculties of Instance-Based Learning Theory, Theory of Mind, and Transfer of Learning. This model functions by learning from both roles in a security scenario: defender and attacker, and by making predictions of the opponent's beliefs, intentions, and actions. The proposed model can better defend against attacks from a wide range of opponents compared to alternatives that attempt to perform optimally without accounting for human biases. Additionally, the proposed model performs better against a range of human-like behavior by explicitly modeling human transfer of learning, which has not yet been applied to cyber defense scenarios. Results from simulation experiments demonstrate the potential usefulness of cognitively inspired models of agents trained in attack and defense roles and how these insights could potentially be used in real-world cybersecurity.
\end{abstract}


\section{Introduction}
Automated defense systems have historically been designed under the assumption that attacking agents make optimal decisions \cite{sinhaStackelberg}. However, applying these methods to the defense against real human attackers could lead to vulnerabilities \cite{NguyenAnalyzing}. One source of these vulnerabilities is the fact that humans rarely stick to theoretically optimal decision patterns, instead updating their behavior dynamically based on the behavior of their opponent \cite{cooneyLearning}.  Recent methods have attempted to improve the functioning of cyber defense systems by learning to defend against more dynamic human-like behavior \cite{cooneyWarningTime}.

An important source of the dynamic change in behavior of human operators in cyber defense scenarios is that they often switch between multiple related tasks \cite{gutzwillerHumanFactors}. This raises the importance of understanding how human \textit{transfer of learning} impacts dynamic decision making in cyber defense scenarios. Transfer of learning involves improving the learning of new tasks through the application of experience gained in related tasks. The goal of this work is to see how human decision makers dynamically adapt their behavior by applying transfer of learning.

To achieve this goal, we will use the simple multi-player decision environment of Stackelberg Security Games (SSG), which has previously been applied to human decision making in cyber defense scenarios \cite{abbasiAdversaries}. SSG is a simple game between two agents, an attacker and a defender, making sequential decisions that determine the reward payoff for both players. Although these games are very simple, they have the potential for broad implications and real-world applications in cybersecurity, infrastructure security, and opportunistic crime, among others \cite{fangSecurity,sinhaStackelberg}.

Existing real-world applications of SSGs often involve protecting against attacks or misuse by humans \cite{fangDeploying}, which motivates the use of models that account for human behavior. One approach to training agents to better respond to real-world environments with human attackers is to learn relative to models of human behavior that are \textit{boundedly rational} \cite{NguyenAnalyzing, cranfordTowardPersonalized}.

Bounded rationality has a long history in cognitive science and economic decision making as a method of understanding optimal decision making relative to access to information and cognitive capacities of individuals \cite{simonBehavioral}. Previous approaches to accounting for the realities of human decision making in SSGs have done so through the application of boundedly rational analysis \cite{sinhaStackelberg,wuInverse}.

These previous methods for improving performance against human attackers have focused on how bounded rationality introduces systematic biases in human decision making. Relatively little attention has been paid to how humans overcome these limitations. In many real-world environments, human decision makers can improve their performance by gaining experience making the decisions of their opponent \cite{gonzalezLearning}. This connects the cyber defense scenario to human transfer of learning, which humans can use to overcome the biases introduced by the bounds of their rationality.

To account for how humans overcome their biases and constraints in cybersecurity environments,
we propose a model that can improve defense by transferring experience in the role of the attacker to the defense side. This method is inspired by \textit{theory of mind} (ToM), by which humans are able to predict the beliefs, desires, and behaviors of other humans \cite{nguyenTheory}. Without taking into account the ToM reasoning that humans use in cyber defense scenarios, existing approaches are less equipped to handle human attacks. Using ToM to improve transfer of learning in a defense scenario involves predicting the attacker's behavior based on experience as an attacker in similar environments.

The model presented in this work is based on Instance-Based Learning Theory (IBLT), which models cognitive agents while accounting for their limitations and cognitive capacities \cite{gonzalezInstance}. A cognitive model based on IBLT, referred to as an Instance-Based Learning (IBL) model, is adjusted to allow for predictions of other agent's behavior, which can dynamically adapt to behavioral biases and strategies of opponents. The result is a model inspired by both bounded rationality and ToM. Experimental results comparing this model with a strategy that attempts to optimally learn to maximize utility, the upper confidence bound model, demonstrates the benefit of the proposed model. Additional comparison is done against another model that is similarly inspired by human cognition, but without the ToM reasoning capabilities.

These experimentation results demonstrate the usefulness of incorporating ToM reasoning in defense models that attempt to perform better against real human decision makers. While the results presented here use simulations, this has the benefit of allowing for a comparison against a wide range of opponents. Future work that draws on these results can focus on comparing the performance of the proposed model when controlling a defense system against decisions made by humans.

\section{Background and Related Work}
\subsection{Stackelberg Security Games}
In SSGs the defending agent is tasked with protecting a number of assets using a limited amount of resources. The defending agent takes the first move of the game by choosing which target or targets to cover or defend. Then, the attacking agent chooses a target or targets to attack. For an asset $t \in T$, where $T$ is the set of all assets, the utility of the defending agent is $U_d^c(t)$ if it is covered and $U_d^u(t)$ if it is uncovered. The attacker's utility is similarly defined as $U_a^c(t)$ if it is covered and $U_a^u(t)$ \cite{korzhykStackelberg}.

Typically, attacking uncovered targets results in a higher utility, and it is the goal of the attacker to select uncovered assets. The defender's goal is to cover the assets that will be selected by the attacker. These four values are defined for each target in the game, which results in a description of the dynamics of the environment.

The traditional formulation of SSGs functions under the assumption that the attacking agent optimally determines its behavior in response to observations of the strategy of the defending player \cite{korzhykStackelberg}. However, more recent applications in boundedly rational decision making in SSGs have relaxed this assumption \cite{fangDeploying}. This is reasonable given the imperfect behavior of real human decision makers, who are biased by cognitive constraints and previous experience.

This describes the one-step version of a SSG, which will be used as a model comparison environment in this paper. Many alternative formulations of SSGs exist, such as a two-stage version that includes an additional step of the defender choosing whether to signal that the asset is covered, and the attacker confirming the attack based on this. We utilize the simplest description of the SSG to allow for more in-depth analysis of the differences between models' behavior.

\subsection{Instance-Based Learning}
In IBL models, agents memory consists of instances $(s,a,x)$ defined by the state $s$, their action $a$, and the outcome $x$ \cite{gonzalezInstance}. These are stored in memory as outcomes $x$ and options $k = (s,a)$. At time $t$ there may be $n_{k,t}$ generated instances $(k,x_{i,k,t})$. Agents take actions that maximize expected utility calculated through the blending function according to the equation:
\begin{equation}
V_{k,t} = \sum_{i=1}^{n_{k,t}}p_{i,k,t} x_{i,k,t}
\end{equation}
where $x$ are the outcomes, and the probability of retrieval is $p_{i,k,t}$. This probability of retrieval is calculated as:
\begin{equation}
p_{i,k,t} = \dfrac{\exp{(\Lambda_{i,k,t}/\tau)}}{\sum_{j=1}^{n_{k,t}}\exp{(\Lambda_{j,k,t}/\tau)}}
\end{equation}
where $\tau$ is a temperature parameter and the activation $\Lambda_{i,k,t}$ value, which represents the ease of recalling a specific instance in memory, calculated according to the recency through the equation:

\begin{equation}
\Lambda_{i,k,t} = \ln \bigg( \sum_{t' \in T_{i,k,t}}(t - t')^{-d}\bigg) + \sigma \ln \dfrac{1 - \xi_{i,k,t}}{\xi_{i,k,t}}
\end{equation}
where $d$ and $\sigma$ are decay and noise parameters, and $T_{i,k,t} \subset \{0,...,t-1\}$ is the previous observations of the instance $i$. The term $\xi_{i,k,t}$ is a random number drawn from a unit uniform distribution, which introduces noise to account for individual differences in recall. Because of the relationship between noise $\sigma$ and temperature $\tau$ in IBLT, the temperature parameter $\tau$ is typically set to $\sigma \sqrt{2}$.  

\subsection{Bounded Rationality}
IBL models can incorporate the motivations of bounded rationality through the decay $d$ and noise $\sigma$ parameters. These parameters influence the activation function of an instance and, in turn, the probability of that instance being retrieved from memory and incorporated into the expected utility calculation. Agents with less cognitive resources may forget instances more quickly due to faster decay, or act more stochastically as a result of higher noise, and as a result impact the evaluation of expected utility.

One common feature of boundedly rational models of human decision making is the use of probability distributions on possible actions, instead of deterministic action selection \cite{mattssonProbabilistic}. This approach can have higher predictive accuracy over deterministic selection and reflect the stochastic nature of human decision making. This type of stochastic probability distribution can be calculated in IBL models through the use of a soft-max over the expected utilities $V_{k,t}$ of each option available to an agent according to:
\begin{equation}
p(k_i) = \dfrac{\exp{(V_{i,k_i,t}/\tau_v)}}{\sum_{k_j=k_1}^{k_n}\exp{(V_{i,k_j,t}/\tau_v)}} \label{eq:softmax}
\end{equation}
where $\tau_v$ is a temperature parameter controlling the likelihood that the distribution favors the option with a higher estimated utility $V_{i,k_i,t}$. This equation calculates a probability $p(k)$ of selecting the option $k$ out of the set of possible options $k = \{k_o, ... , k_n\}$.

\begin{figure*}
  \centering
  \includegraphics[width=0.9\textwidth]{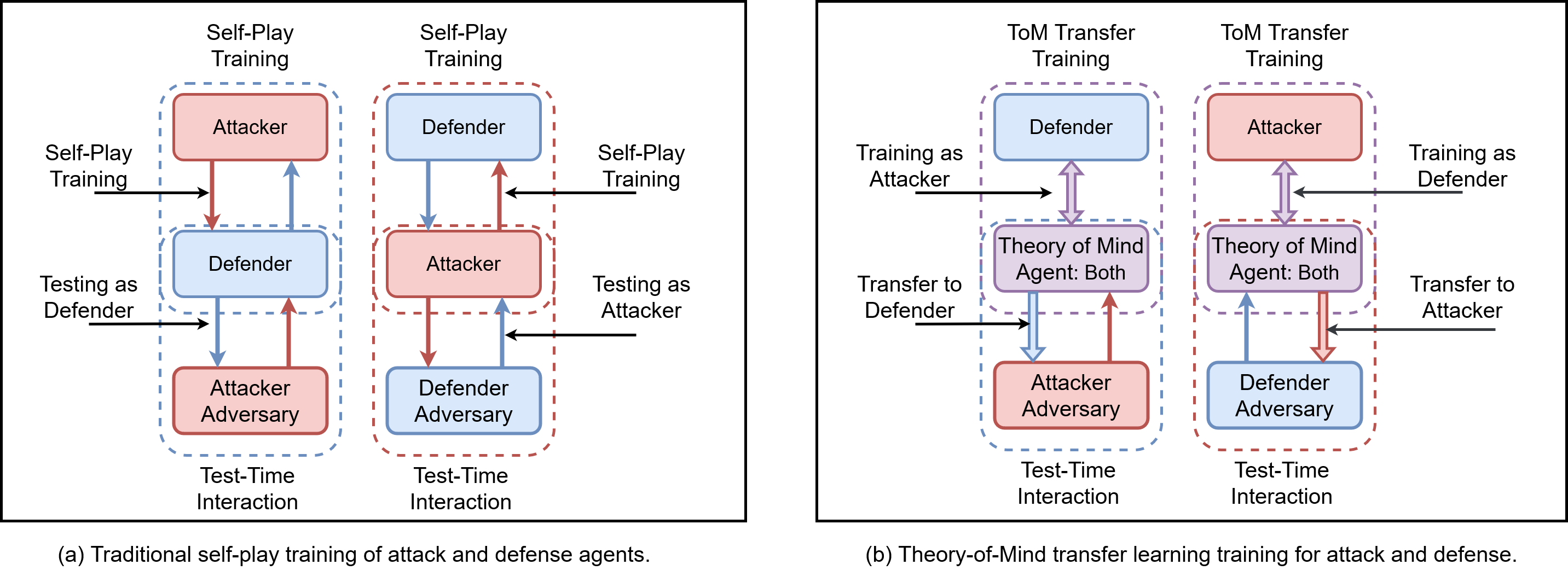}
  \caption{Comparison of (a) self-play training for attack and defense agents  and (b) our proposed theory-of-mind transfer of learning training .}
  \label{fig:TransferLearning}
\end{figure*}

\subsection{Theory of Mind}
Previous methods for introducing ToM to cognitive models inspired by IBLT have demonstrated more human-like predictions of agent behavior \cite{nguyenTheory}. This is an important feature of models used in security games with real-world applications, as the methods that humans use to predict the behavior and beliefs of other agents are highly relevant to these applications.

An approach to human-like ToM in artificial intelligence models is through the use of inverse reinforcement learning (RL) \cite{jaraEttingerTheory}. The method of achieving ToM modeling in IBLT presented in this work is analogous to inverse RL applied to IBL. While these methods have demonstrated impressive success, these types of RL algorithm applications are not fundamentally grounded in cognitive theory in the same way that IBLT is.

Specifically, RL methods typically focus on optimizing performance, rather than capturing the biases and constraints present in human learning \cite{suttonReinforcement}. In the experimentation portion of this paper, we compare the performance of our proposed method against a closely related approach to RL, the equations of the upper confidence bound \cite{suttonReinforcement}.  

\subsection{Transfer of Learning}
Human bounded rationality introduces potential biases that can affect performance in a variety of tasks. These biases can affect generalization and are the result of biological factors that cannot be easily overcome \cite{simsEfficient}. The result of these biases has the potential to significantly limit the transfer of previous experience to a new learning environment. In multi-agent environments, slow learning and biased decision making can lead to behavior that fails to collaborate in cooperative tasks, and is easily exploited in competitive tasks. One way in which humans are able to overcome these difficulties is through the use of transfer of learning \cite{blickensderferCross}.

In cognitive science and artificial intelligence research, ``transfer of learning" refers to a group of methods used to improve performance in one domain by transferring learned experience from a related domain (for review see \cite{weissSurvey}). This concept is closely related to \textit{cross-training} within the field of cognitive science, whereby human learners are trained in multiple interacting roles, which has been shown to improve performance in their respective roles \cite{blickensderferCross}.

Humans have a natural ability to improve performance in team and competitive environments by gaining experience in the role of their teammates or competitors \cite{gonzalezLearning}. This ability is in part due to humans' ability to use ToM reasoning to predict the beliefs, goals, and intentions of others. When humans gain experience in the roles of teammates or competitors, they gain a richer ToM that allows them to better predict the behavior of others, improving collaboration or skill in competition.  

Figure \ref{fig:TransferLearning} details the proposed ToM-inspired transfer of learning paradigm for attack and defense scenarios. This is compared to the traditional self-play training design, in which the agent's roles are more clearly delimited. This method of training to make predictions of opponent behavior using the ToM and transferring experience into the alternative role has the potential to improve performance.

\subsection{Related Models}
One approach related to this work is the Boundedly Rational Quantile Response (BRQR) model which seeks to model human bounded rationality in SSGs through the introduction of a parameter $\lambda$ that represents the rationality of a decision maker \cite{yangImproving}. This parameter can be tuned to the behavior of a human participant and account for their deviation from the utility-maximizing QR model.

Another model that attempts to account for cognitive constraints by modifying the QR algorithm is the Subjective Expected Utility QR (SUQR) model \cite{NguyenAnalyzing}. This method proposes the use of a subjective utility function of the adversary for SSGs according to the equation $\hat{U}_t^a = w_1x_t + w_2R_t^a + w_3P_t^a$. Where $(R_t^a, P_t^a)$ are the attacker's reward and penalty, and $(w_1,w_2,w_3)$ are learnable model parameters.

While these models can produce behavior that matches human performance, they do it through assuming a specific bias from optimal performance in SSGs, and fitting a parameter based on that bias. As opposed to these previous methods, the model proposed in this work is inspired by a cognitive modeling theory that describes general decision making in dynamic environments. The next section describes this proposed model in more detail and how it is applied onto the SSG environment.

The BRQR and SUQR models have previously been compared with a standard IBL model in the context of security games, demonstrating a better fit to human performance through the cognitively inspired IBL model \cite{blickensderferCross}. The model presented in this work differs from standard IBL in that it is altered to make predictions of other agent's beliefs, goals, and behavior. Leveraging ToM affords not only improvements in the role of attacker or defender but significant improvements in the transfer of skills from one role to the other.

While these previous models are related to the model proposed in this work, their differences in motivation are significant enough that comparison in SSGs is unnecessary. This is partially due to the main focus of this work being in assessing the use-fullness of ToM applied to cognitive models trained in SSGs. It would be possible to adapt these previous methods to additionally engage with ToM reasoning. The section following the description of the proposed model will detail the two models used for comparison against the proposed model, a basic version of the IBL model without ToM reasoning, and a model trained to learn tasks similar to SSGs as efficiently as possible. 

\section{Proposed Model}
When applying theory-of-mind modeling onto IBL, agents additionally keep in their memory predictions of opponent actions $a_o$, and of opponent observations $s_o$, if this is not directly observed themselves. These values are used to train a model that predicts both the behavior of other agents and a policy that is useful if their role in the game switches. The result of this theory-of-mind inspired prediction of other agent's beliefs and/or actions is the proposed Instance-Based Theory of Mind (IBToM) model. These opponent options $k_o = (s_o,a_o)$ are used similarly as with the standard IBL model, except that they predict the opponent's behavior. This is done through an expected utility function:
\begin{equation}
O_{k_o,t} = \sum_{i=1}^{n_{k_o,t}}  p_{i,k_o,t} x_o
\end{equation}
Where $x_o$ is the observed opponent outcome after selecting option $k_o$. This results in an expected utility function that assigns higher probability to actions that match opponent's previous behavior, additionally accounting for the probability of retrieval and activation function that form the basis of IBL models. When the opponent's outcome is unknown, this can be replaced by an identity function $\mathbbm{1}_{i,k_o,t}$ equal to 1 for the options selected by the opponent and 0 otherwise.

The proposed model is also able to make predictions of opponent stochastic behavior in a similar manner as described in Eq \ref{eq:softmax}.
\begin{equation}
	\label{eq:oppExpectedUtility}
	p(k_o) = \dfrac{\exp(O_{i,k_o,t} / \tau_o)}{\sum_{k_j=k_1}^{b_n} \exp(O_{i,k_j,t} / \tau_o)}
\end{equation}
This makes the proposed model well equipped to make predictions of human opponents and team-mates in complex environments that require ToM reasoning.

\begin{algorithm}[t]
\caption{Instance-Based Theory of Mind Algorithm}\label{alg:cap}
\begin{algorithmic}
\State $M \leftarrow \{ \}$
\While{$t < \text{episode length}$}
\State $\text{sample } k_o \sim p(k_0) \leftarrow  \dfrac{\exp(O_{i,k_o,t} / \tau_o)}{\sum_{k_j=k_1}^{b_n} \exp(O_{i,k_j,t} / \tau_o)}$\\
\State $k_i \leftarrow (s,a_i,k_o) \text{  } \forall_{a_i}$ \\
\State $\text{sample } k \sim p(k_i) \leftarrow  \dfrac{\exp(V_{i,k_i,t} / \tau_v)}{\sum_{k_j=k_1}^{b_n} \exp(O_{i,k_j,t} / \tau_v)}$\\
\State $x \leftarrow p(x|k)$
\State $\text{Perform option } k \text{ and observe outcome } x$
\State $\text{Observe opponent option and outcome: } k^*_o \text{ } x_o$
\State $M \leftarrow \{M \cup \{s,a,x,s_o,a^*_o,x_o\} \}$
\EndWhile
\end{algorithmic}
\end{algorithm}

Algorithm \ref{alg:cap} details the functioning of the IBToM algorithm, which can be applied onto predicting opponent behavior to improve performance in the SSG environments used in model comparison, as well as transfer of learning in the same domain. This algorithm functions by predicting the selection of opponent options $k_o$ by sampling the probability distribution defined in Equation \ref{eq:oppExpectedUtility}.

\begin{figure*}[!ht]
  \centering
  \includegraphics[width=\textwidth ]{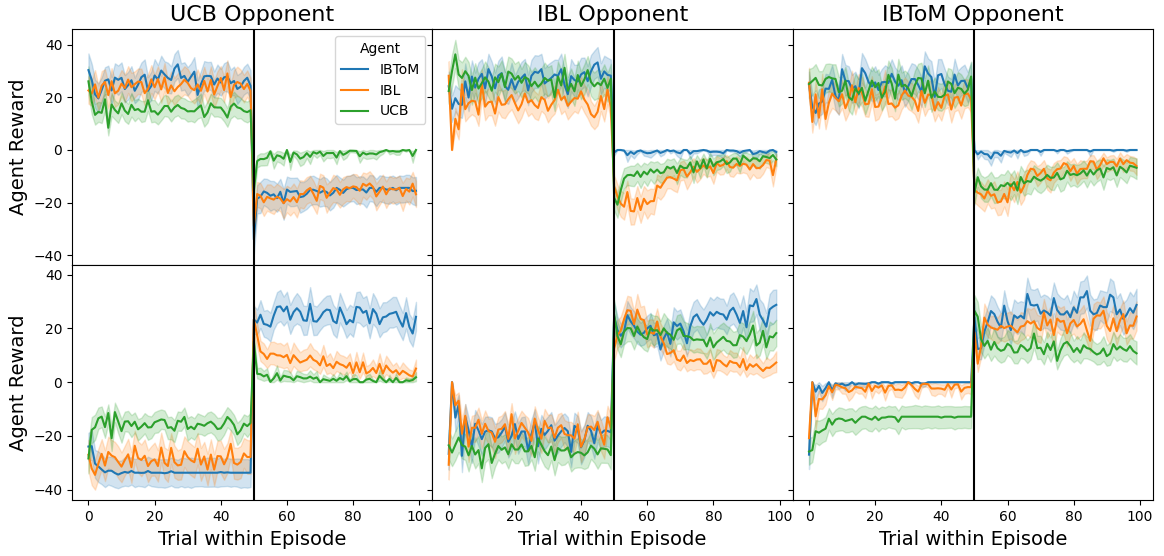}
  \caption{\textbf{Top:} Agent reward when paired against Upper Confidence Bound (left), Instance-Based Learning (middle) and Instance-Based Theory of Mind (right). First 50 trials are as the attacking agent, then 50 trials as the defending agent. \textbf{Bottom:} The same comparison of agent reward paired with other models with reversed training. The first 50 trials are for the defending agent, then 50 trials for the attacking agent. }
  \label{fig:Results}
\end{figure*}

This predicted option selection is used to augment each available option $k_i \leftarrow (s,a_i,k_o)$, so that option selection is made based on predicted opponent behavior. After performing the option $k$ sampled from the probability distribution defined in Equation \ref{eq:softmax}, the agent stores in memory the observed outcome $x$, the observed opponent option that was actually selected $k_o^*$, and the observed opponent outcome $x_o$.

Importantly, the action stored in memory for the opponent's action is the observed action $a^*_o$, instead of the predicted action $a_o$. If the true observation of opponents is observed after taking an action, then that can similarly be stored in memory as $s^*_o$. The proposed IBToM model uses predictions of opponent actions as features of the state $s$ that is used to calculate the expected utility of an option $k = (s,a)$. This also provides an expected utility function that can be applied directly onto the task of the opponent if the roles of the environment were to switch.
\section{Experimentation}
\subsection{Comparison Models}
Models used for comparison consist of our proposed Instance-Based Theory of Mind (IBToM) model, an Instance-Based Learning (IBL) model without ToM reasoning, and the Upper Confidence Bound (UCB) model. The three models used for comparison take on the role of both attacker and defender. 1000 pairs of each model combination of models are trained.

Aside from the standard IBL model, the main model for comparison in the experiments used to evaluate the model proposed in the following section is based on the Upper Confidence Bound algorithm \cite{suttonReinforcement}, which selects actions $A_t$ according to the maximization:
\begin{equation}
A_t = \text{argmax}_a \bigg[ Q_t(a) + c \sqrt{\dfrac{\ln t}{N_t(a)}} \bigg]
\end{equation}
where $N_t(a)$ is the number of times the action $a$ has been taken at time $t$, $Q_t(a)$ is the mean value of the action $a$, and $c$ is a parameter that controls the level of confidence in how previous actions should dictate future ones. This algorithm produces actions that minimize the regret observed in stateless decision tasks such as SSGs \cite{suttonReinforcement}.

\begin{table}[h]
\begin{center}
\begin{tabular}{||c c c c c||}
 \hline
 Model & Inv. Temp & Noise & Decay & Exploration \\ [0.3ex]
 \hline\hline
 IBToM & 0.05 & 0.25 & 0.5 & N/A \\
 \hline
 IBL & 0.05 & 0.25 & 0.5 & N/A \\
 \hline
 UCB & 0.05 & N/A & N/A & 10 \\
 \hline
 \hline
\end{tabular}
\caption{\label{tab:parameters}Model comparison parameters.}
\end{center}
\end{table}

All model parameters (Table \ref{tab:parameters}) are selected from standard baselines described in the literature of IBLT \cite{gonzalezInstance} and methods related to reinforcement learning \cite{suttonReinforcement}, relative to the average utility of the two assets in the SSG.

\subsection{Task and Results}
A single-step Stackelberg security game with two assets is used as an experimentation environment for model comparison. In this task, the defending agent selects which asset it will protect before the agent selects which asset it will attack. If the attacking agent selects the defended asset, both agents receive no reward. If the attacking agent selects the undefended action, the attacker receives a positive reward equal to the value of the asset, and the defending agent receives a negative reward of the same magnitude. This results in a zero-sum style Stackelberg security game with variable asset values.

Assets values are determined by a Dirichlet($[3,4]$) distribution multiplied by 100, meaning the values between the two boxes summed to 100 and typically ranged from roughly 25-75. This distribution was sampled from at the beginning of each episode. Each model was reset at the beginning of each episode, so that decisions were made only based on their experience within a single episode.

The main metric of model comparison is the reward observed by the training agent, either attacking or defending. If the attacking agent successfully accessed an asset, they received a reward equal to its value, and the defender lost the equivalent reward. If the defending agent successfully protected the asset, both agents received zero reward. Model pairs were trained together for 50 trials choosing the decisions of either the attacker or defender. After these initial 50 trials, agents switched roles to the alternative position for another 50 trials.

The top row results shown in Figure \ref{fig:Results} compare the performance of each of the three models under comparison (UCB, IBL, and IBToM) when paired against an opponent of the type indicated by the row of the figure (UCB, IBL, and IBToM). For the first 50 trials of an episode, the agent performs the actions of the defending agent against their opponent. Then, the agent's policies are inverted so that they correspond to the alternative role in the game, for the IBL and UCB models. For the IBToM model, the opponent prediction model is used when switching roles. These results indicate that the IBToM model effectively learns to predict the behavior of opponents in the SSG environment, and is able to use those predictions to inform decisions when the role of the task is switched.

In addition to testing the average reward of models against the opponents they were paired with during training, we were also interested in how the end of training behavior of each model performed when paired against models they did not train with. This type of out of training distribution opponent performance is a significant problem for models in both cooperative and competitive environments.

To assess this, we tested models using the same parameters (Table \ref{tab:parameters}) against a sampling of 1000 versions of each model with randomized parameters. Parameters were randomized by sampling a Beta(10,10) distribution shifted by the original default parameter values used in training. This parameter randomization is done to reflect the natural deviation in opponent strategies that exists in real world settings. 

\begin{figure}[t]
  \centering
  \includegraphics[width=\columnwidth ]{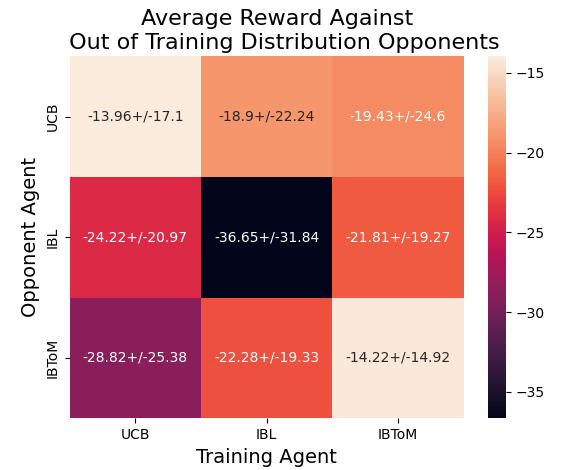}
  \caption{Performance of trained models making the decisions of the defender agent against a population of all models with randomized parameters when selecting the actions of the attacker.}
  \label{tab:Results2}
\end{figure}
The results shown in Figure \ref{tab:Results2} demonstrate that the proposed IBToM model has significantly (Tukey's HSD Pairwise Group Comparisons: $p<0.001$) higher average episode reward against a random sampling of opponents. This indicates that the ToM training has useful implications for real world applications of SSGs, since they are more likely to encounter scenarios with agents who learn and behave differently than the agents they are initially trained with. Interestingly, while the IBToM model had the highest average performance against out of distribution opponents of all types, the IBL model performed slightly better against the random sampling of IBL opponents.

We also are interested in the performance of the proposed model in simulating attacker behaviour after gaining experience as a defender. While cyber defense scenarios are typically investigated from the perspective of the defending agent, designing attacking agents that perform as well as possible can allow for better comparison of defense systems. 

\begin{figure}[t]
  \centering
  \includegraphics[width=\columnwidth ]{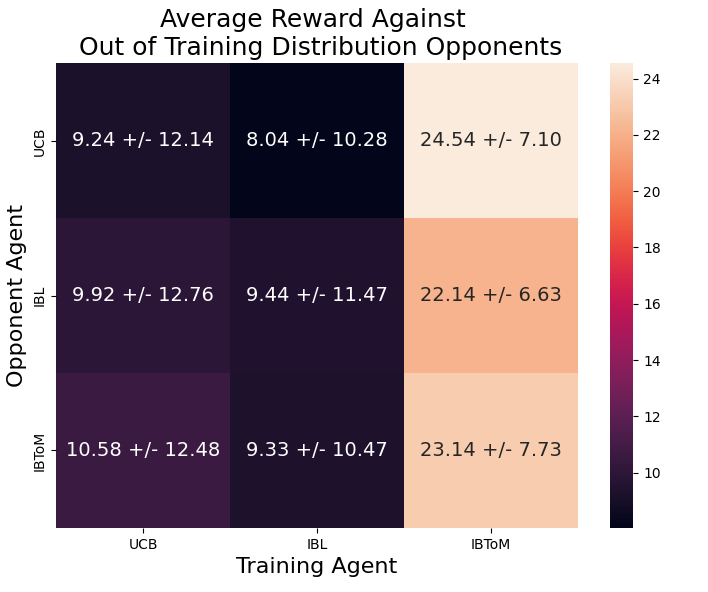}
  \caption{Performance of trained models making the decisions of the attacker agent against a population of all models with randomized parameters when selecting the actions of the defender.}
  \label{tab:Results3}
\end{figure}

The results in Figure \ref{tab:Results3} demonstrate that the proposed IBToM model has the highest performance when choosing the decisions of the attacker agent when paired against a population of randomized defenders. This demonstrates that the IBToM model is not only a useful tool for designing defense systems, but also evaluating alternative defense systems. 

\section{Conclusions and Future Directions}
Results from experimentation in a simple Stackelberg Security Game indicate that theory of mind can improve transfer of learning in cognitive models. The model presented in this work applied theory of mind modeling to improve boundedly-rational behavior in a competitive environment that simulated a simple security game. These results further support the application of such cognitive models as a proxy of human learning in a variety of tasks.

Using theory of mind as a means of overcoming bounded rationality is also of interest in cooperative games where parts of the environment are not observed by all players, such as the card game Hanabi \cite{malloyGeneralization}. Modeling teammate and opponent strategies as boundedly-rational has also demonstrated success in mixed cooperative and competitive environments \cite{malloyCapacity}.

The model we present involved a simple version of theory of mind, predicting only the actions of other agents and not their observations. This was appropriate for the simplicity of the SSG design used in experimentation, but further additions to this model are possible. The most notable method for enhancing this model's theory of mind is through predictions of opponent observations for environments that do not share observations across agents. This could be done in the SSG environment, by adding features to assets that indicate their value.

For improving transfer of learning through theory of mind reasoning and applying experience as an opponent, this type of belief and unobserved state prediction demonstrates a high potential for improved performance. Ideally, gaining experience as an opponent or collaborator while applying theory of mind reasoning would significantly improve the applicability of ToM reasoning onto transfer of learning. This has the potential to improve transfer of learning as well as performance in some of the most difficult tasks, where agents do not observe the whole environment or the observations of other agents.

\section*{Acknowledgements}
This research was sponsored by the Army Research Office and accomplished under Australia-US MURI Grant Number W911NF-20-S-000 and by the Army Research Laboratory under Cooperative Agreement Number W911NF-13-2-0045 (ARL Cyber Security CRA)

\end{document}